\documentclass[runningheads]{llncs}
\usepackage{graphicx}
\begin{document}
\title{Adapting the Human: Leveraging Wearable Technology in HRI \thanks{This work has been supported from the European Union’s Horizon  2020  research  and  innovation  programme  under grant  agreement  No  688117  “Safe  human-robot  interaction  in  logistic  applications  for  highly  flexible  warehouses(SafeLog)”.}}
%
%
\author{David Puljiz\inst{1}\orcidID{0000-0002-8966-3091} \and
Bj\"orn Hein\inst{1,2}\orcidID{0000-0001-9569-5201}}
\authorrunning{D. Puljiz et al.}
%
\institute{Intelligent Process Automation and Robotics Lab (IPR), Institute for Anthropomatics and Robotics, Karlsruhe Institute of Technology, Karlsruhe, Germany
\email{\{david.puljiz,bjoern.hein\}@kit.edu}\\ \and
Karlsruhe University of Applied Sciences, Karlsruhe, Germany
}
\maketitle              
\begin{abstract}
Adhering to current HRI paradigms, all of the sensors, visualisation and legibility of actions and motions are borne by the robot or its working cell. This necessarily makes robots more complex or confines them into specialised, structured environments. We propose leveraging the state of the art of wearable technologies, such as augmented reality head mounted displays, smart watches, sensor tags and radio-frequency ranging, to "adapt" the human and reduce the requirements and complexity of robots.     

\end{abstract}

Every robot that interacts with humans needs to fulfil many basic requirements, no matter if in a social, service or industrial context - compliance, either passive mechanical or active through force-torque sensors; safety through laser scanners, cameras etc.; legibility of motions and actions, through more human-like motions or by adding other visual cues; human intention estimation to facilitate cooperation, and many more. All of these sensors, visualisations and legibility of actions and motions are borne by the robot or its working cell. This makes robots that interact with humans complex and costly. \par

One possible alternative could be to meet the robots half-way and "adapt" the humans too, so the robots have an easier time fulfilling the requirements without increasing complexity. The human would carry all or part of the visualisation systems and sensors needed for an intuitive interaction. These would come in the form of wearables - augmented reality (AR) head-mounted displays (HMDs) such as the HoloLens, MagicLeap etc., sensor tags or smart watches, and specialised radio frequency (RF) ranging units for functional safety. Except for the ranging components the presented wearables are already used, or will be in the near future, in everyday life, thus not presenting any extra burden for the human. This trend will only increase as newer technologies, such as smart textiles, become ever more present. Thus, leveraging wearables should become a key point in future HRI paradigms.\par 

The wearables here would act as a translators from human-interpretable data to robot-interpretable data and vice versa, meaning that any interaction between a human and a robot can then be boiled down to machine-to-machine communication. \par

The HMDs present the central component in such a system as they are responsible for audio and AR cues, carry most of the input modalities and posses almost all of the sensors. In several works we have shown that HMDs allow easier programming of robots \cite{puljiz2019hand}, display of robot intentions \cite{puljiz2018impl}, estimation of intentions \cite{PETKOVIC2019hir} and tracking of the human coworker  \cite{puljiz2019concepts} etc. This makes the interaction more intuitive for the human and easier to understand for the robot. More information means that the robotic system has to deal with less uncertainty, making decisions simpler. Although HMDs are confined to enterprises at the moment, they are expect to be available for personal use in the next few years. \par   

Another interesting use of AR in social robotics in particular comes from Hirokazu Kato during the 2nd VAM-HRI workshop in 2019 \cite{williams2019vamhri}. He proposed that the agent the human interacts with is disembodied. This agent could then "inhabit" different robots and systems or be disembodied when the system cannot be present (e.g. a Roomba cannot climb stairs). Agents can of course be personalised. This particular modality may be further enhanced with cloud robotics. \par

Smart watches with inertial measurement units (IMUs) and/or sensor tags are used to track a person's arms when outside the field of view of the HMDs sensors. This is not strictly necessary but enhances safety and makes gesture commands more intuitive as they are not restricted by the sensors' fields of view. The IMUs accumulate errors fast, thus they still need the sensors of the HMD to correct them and are not a stand-alone solution. \par    

In the European project SafeLog one of the main challenges is ensuring human safety inside automated warehouses. Adapting the mobile robot to be safe was not an option as modifying fleets larger than 500 robots would be financially unfeasible. The answer was to outfit the robots with simple and cheap hardware, and let the human wear the complex part of the safety system, based on RF ranging \cite{ivsic2020safelog}. As the robots would largely outnumber the humans in the warehouse, such a system would be much cheaper than e.g. outfitting the robots with laser sensors. Though the system proved highly effective in scenarios where the robots significantly outnumber the humans, it also reduces the number of sensor and complexity of the robots themselves. Thus it is applicable in all manner of HRI scenarios. \par

The vision of such a worn systems is purely the realm of research at the moment, however the enabling technologies are already there or soon to be. Such an approach could be enhanced in the future with heart rate sensors to estimate human emotions \cite{itoh2006hri}, electromyography sensors for better action estimation \cite{assad2013biosleeve} etc. We must move away from our anthropocentric view of HRI, where we expect the robot to do everything, and move towards paradigms where humans too wear sensors and hardware to better translate the human world to the robot world and improve cooperation between them. \par 

\bibliographystyle{ieeetr}
\bibliography{references.bib}
\end{document}